\pdfoutput=1

\documentclass[11pt]{article}

\usepackage[]{acl}

\usepackage{times}
\usepackage{latexsym}

\usepackage{amsmath}
\usepackage{amsfonts}
\usepackage{multirow}
\usepackage{booktabs}
\usepackage{arydshln}
\usepackage{amssymb}
\usepackage{tikz}
\usepackage{booktabs}
\usepackage{hyperref}
\usepackage{cleveref}
\usepackage{adjustbox}
\usepackage{pgfplots}
\usepackage{subfig}
\usepackage{worldflags}
\usepackage{siunitx}
\usepackage{relsize}
\usepackage{array}
\usepackage{xcolor}
\definecolor{darklavender}{rgb}{0.45, 0.31, 0.59}
\usepackage[normalem]{ulem}
\robustify\uline

\usepackage{color, colortbl}
\usepackage{comment}

\usepackage[T1]{fontenc}

\usepackage[utf8]{inputenc}

\usepackage{microtype}

\usepackage{inconsolata}

\usepackage{graphicx}

\title{Domain-specific or Uncertainty-aware models: Does it \textit{really} make a difference for \textcolor{black}{biomedical text classification}? }

\author{
 \textbf{Aman Sinha\textsuperscript{$\clubsuit$,$\heartsuit$}},
 \textbf{Timothee Mickus\textsuperscript{$\spadesuit$}},
 \textbf{Marianne Clausel\textsuperscript{$\clubsuit$}}, \\
 \textbf{Mathieu Constant\textsuperscript{$\clubsuit$}} and \textbf{Xavier Coubez\textsuperscript{$\heartsuit$}}
\\
\\
 \textsuperscript{$\clubsuit$}Université de Lorraine, Nancy, France \\
 \textsuperscript{$\spadesuit$}University of Helsinki, Helsinki, Finland \\
 \textsuperscript{$\heartsuit$}Institut de Cancérologie, Strasbourg, France
\\
 \small{
   \textbf{Correspondence:} \href{mailto:aman.sinha@univ-lorraine.fr}{\tt aman.sinha@univ-lorraine.fr}
 }
}

\begin{document}
\maketitle
\begin{abstract}

{\color{black}
The success of pretrained language models (PLMs) across a spate of use-cases has led to significant investment from the NLP community towards building domain-specific foundational models.
On the other hand, in mission critical settings such as biomedical applications, other aspects also factor in---chief of which is a model's ability to produce reasonable estimates of its own uncertainty.
}
In the present study, we discuss these two desiderata through the lens of how they shape the entropy of a model's output probability distribution. 
We find that domain specificity and uncertainty awareness can often be successfully combined\textcolor{black}{, but the exact task at hand weighs in much more strongly.
}
\end{abstract}

\section{Introduction}
Deep-learning models are trained with data-driven approaches to maximize prediction accuracy \cite{goodfellow2016deep}. 
This entails several well-documented pitfalls, ranging from closed-domain limitations \cite{daume2006domain} to social systemic biases \cite{mccoy2019right, schnabel2016recommendations}.
These limitations compound to a severe deterioration of model performances in out-of-domain (OOD) scenarios \cite{ hurd2013monetary, shah2020pitfalls}.
This has led to engineering efforts towards developing models tailored to specific domains, ranging from the legal \citep{10.1145/3594536.3595165} to the biomedical \citep{lee2020biobert, Singhal2023TowardsEM} ones.


Domain-specific models, while useful, are rarely considered as a definitive answer. 
Crucially, in the biomedical domain, experts require more reliability from these models---in particular, insofar as accounting for uncertainty in prediction is concerned. 
For example, in the case of a risk scoring model used to rank patients for live transplant, uncertainty-awareness becomes critical. The lack of uncertainty-aware models may lead to improper allocation of medical resources \cite{steyerberg2010assessing}. Such concerns exemplify the  importance of uncertainty aware models and its critical role in model selection.




The compatibility of domain-specific pretraining and uncertainty modeling appears under-assessed. To illustrate this, one can consider the entropy of output distributions: Domain-specific pretraining will lead to more probability mass assigned to a single (hopefully correct) estimate, leading to a lower entropy; 
whereas uncertainty-aware designs intend to not neglect valid alternatives---meaning that the probability mass should be spread out, which entails a higher entropy when uncertainty is due.
 
\begin{figure}[!t]
    \centering
    \resizebox{0.85\columnwidth}{!}{
        \begin{tabular}{>{\centering\arraybackslash}m{0.1\columnwidth} >{\centering\arraybackslash}m{0.5\columnwidth}@{{\quad}} >{\centering\arraybackslash}m{0.5\columnwidth}}
            & {\Large Uncertainty-unaware} & {\Large \bf Uncertainty-aware} \\
        \rotatebox{90}{\parbox{0.25\columnwidth}{\centering \Large General-domain}}
             &
             \begin{tikzpicture}
               \node[color=gray!10, fill, rounded corners=0.5cm, outer sep=3cm] { \includegraphics[width=0.4\columnwidth, trim=0.75cm 0.35cm 0.75cm 0cm, clip]{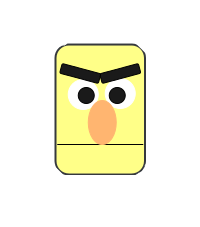} };  
             \end{tikzpicture}
              & 
             \begin{tikzpicture}
               \node[color=blue!15, fill, rounded corners=0.5cm, outer sep=3cm] { \includegraphics[width=0.4\columnwidth, trim=0.75cm 0.35cm 0.75cm 0cm, clip]{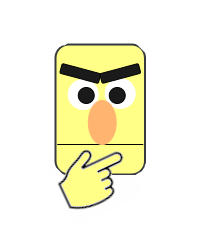} };  
             \end{tikzpicture}\\

         && \\
        \rotatebox{90}{\parbox{0.25\columnwidth}{\centering \Large \bf Domain-specific}}
             & 
             \begin{tikzpicture}
               \node[color=red!15, fill, rounded corners=0.5cm, outer sep=3cm] { \includegraphics[width=0.4\columnwidth, trim=0.75cm 0.35cm 0.75cm 0cm, clip]{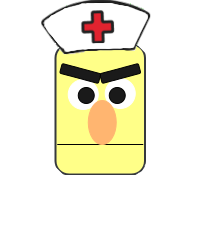} };  
             \end{tikzpicture}
              & 
             \begin{tikzpicture}
               \node[color=darklavender!35, fill, rounded corners=0.5cm, outer sep=3cm] { \includegraphics[width=0.4\columnwidth, trim=0.75cm 0.35cm 0.75cm 0cm, clip]{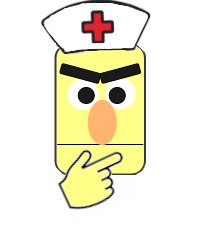} };  
             \end{tikzpicture}\\
        \end{tabular}
    
    }
    
    \caption{Illustration of this study's setup. We perform a systematic comparison of domain-specificity and uncertainty-awareness in the medical domain.}
    \label{fig:intro}
\end{figure}

\begin{table*}[!t]
    \centering
    \resizebox{\textwidth}{!}{
    \begin{tabular}{l@{{~}}l p{11cm}c cc cc cccc}
    \hline
        \multicolumn{2}{c}{\multirow{2}{*}{Dataset}} &
        \multirow{2}{*}{Task Description} & &
        \multicolumn{3}{c}{Splits} & &\multicolumn{4}{c}{Statistics}\\
        \cline{5-7} \cline{9-12}
         & && &  train & val & test &&  \texttt{\#Class} & \texttt{CIR}&\texttt{avglen}  & \texttt{maxlen} \\
    \hline
      \resizebox{2ex}{!}{\worldflag{US}} & MedABS   & Predict the patient condition described,  given a medical abstract  & &8662&2888&2888&&5&  3.1445&180.59&597\\
      \resizebox{2ex}{!}{\worldflag{US}} & MedNLI   & Predict the inference type, given a hypothesis and a premise & &
      11232& 1395&1422 && 3 & 1& 23.83& 151\\
      \resizebox{2ex}{!}{\worldflag{US}} & SMOKING  & Predict the patient smoking status, given a
medical discharge record &&398&100&104&&5 & 23.75 &654.30&2788\\

      \resizebox{2ex}{!}{\worldflag{FR}} & PxSLU  & Predict the drug prescription intent, given a user speech transcription&  & 1386&	198	&397&& 4 & 98.1538& 11.40 & 48\\
      \resizebox{2ex}{!}{\worldflag{FR}} & MedMCQA & Predict the number of answers, given a medical multi-choice question & &  2171&	312	&622&&5	& 21.1176& 12.90 & 92\\
      \resizebox{2ex}{!}{\worldflag{FR}} & MORFITT &  Predict the speciality, given a scientific article abstract & & 1514&	1022&	1088&&12 &15.3529 &226.33 & 1425 \\

    \bottomrule
    \end{tabular}
    }
    \caption{Datasets description. \texttt{CIR} denotes class imbalance ratio. }
    \label{tab:dataset}
\end{table*}
In this work, we reflect on how model-specificity and uncertainty-awareness articulate with one another.
Figure \ref{fig:intro} illustrates the experimental setup we use for our study.  
In practice, we study the performances of frequentist and Bayesian general and domain-specific models on biomedical  text classification tasks across a wide array of metrics, ranging from macro F1 to SCE, with a specific focus on entropy \cite{ruder-plank-2017-learning, kuhn2023semantic}.
More narrowly, we study the following research questions:
\textbf{RQ1}: \textit{Are the benefits of uncertainty-awareness and domain-specificity orthogonal?} 
\textbf{RQ2}: Given our benchmarking results, \textit{should medical practitioners prioritize domain-specificity or uncertainty-awareness?}


\section{Related Work}

Recently, uncertainty quantification has gained attention from the NLP community \cite{xiao2019quantifying, xiao2022uncertainty, hu2023uncertainty}---particularly in mission critical settings, such as in the medical domain \cite{hwang2023uncertainty, barandas2024evaluation}. 
In parallel, compared to domain adaptation approaches \cite{wiese2017neural} for the medical domain, there is a growing interest in domain-specific language models starting from BioBERT \cite{lee2020biobert} to the recent MedPalM \cite{Singhal2023TowardsEM}. 
\textcolor{black}{\citet{xiao2022uncertainty} presented an elaborate study of uncertainty paradigm for \emph{general-domain} PLMs.} 
While uncertainty modeling has been applied to biomedical data previously \citep[e.g.,][]{Begoli2019,ABDAR2021243}, surprisingly little has been done for biomedical textual data.
\textcolor{black}{Therefore, our study precisely focuses on the interaction between the two paradigms for medical domain NLP tasks.}
We address this gap by focusing specifically on predictive entropy \cite{ruder-plank-2017-learning, kuhn2023semantic}.

\section{Methodology}

\paragraph{Datasets.}
We conduct experiments on six standard biomedical datasets: 
three English  datasets, 
viz. MedABS \cite{10.1145/3582768.3582795}, MedNLI \cite{romanov2018lessons} and SMOKING \cite{uzuner2008identifying}; 
as well as three French 
datasets, viz. MORFITT \cite{labrak:hal-04131591}, PxSLU \cite{Kocabiyikoglu2022} and MedMCQA \cite{labrak2023frenchmedmcqa}. 

%
For MEDABS, SMOKING, PxSLU, and MEDMCQA, we do not perform any special preprocessing. 
For MEDMCQA, we perform Task 2, i.e., predicting the number of possible responses (ranging from 1-5) for the input multi choice question. 
For MEDNLI, we concatenate the statement and hypothesis using the \texttt{[SEP]} token and use it as an input converting it to a multi-class task.
For MORFITT, which is originally a multi-label classification task, we use the first label for each sample to convert it to a multi-class problem. The descriptive statistics of these datasets are listed in Table \ref{tab:dataset}, along with class imbalance ratio (CIR; \citealp{yu2022re}). See Appendix \ref{adx:sup datasets} for more information.

\paragraph{Models.}
We derive classifiers from language-specific PLMs: for English datasets, we use \texttt{BERT} \cite{devlin2018bert} and \texttt{BioBERT} \cite{lee2020biobert}; for French, we use \texttt{CamemBERT} \cite{martin2019camembert} and \texttt{CamemBERT-bio} \cite{touchent:hal-04130187}. 
We compare two types of models, frequentist deep learning models (DNN) and Bayesian deep learning models (BNNs). 
The DNN model comprises of a PLM-based encoder, a Dropout unit along with 1-layer classifier. 
The BNN models 
are likewise based on a PLM encoder, \textcolor{black}{along with a Bayesian module applied over the classification layer. We also experimented with MC-dropout models \cite{gal2016dropout}, DropConnect \cite{mobiny2021dropconnect}, and variational inference \cite{blundell2015weight} models. We focus\footnote{
      We justify our focus on DropConnect empirically, as it yielded the highest validation F1 scores on average in our case. See \Cref{adx:details:models,adx:sup results} for details. {\color{black} All main text results for uncertainty-aware classifiers pertain to DropConnect.}
}
on {\color{black} the DropConnect architecture which comprises} a PLM encoder along \textcolor{black}{a DropConnect} dense \textcolor{black}{classification} layer}\textcolor{black}{.} 
 This approach infuses stochasticity into a deterministic model  
{\color{black} by randomly zeroing out classifier weights with probability $1-p$.}
This 
allows us to sample multiple outputs for a given input, thus enabling to aggregate the predictions and to produce estimates of uncertainty. 

For simplicity, we note domain-specific models as $+\mathcal{D}$ (and general models $-\mathcal{D}$); uncertainty aware models are referred to as $+\mathcal{U}$ (with frequentist models noted $-\mathcal{U}$).
We replicate training across 10 seeds per model and dataset; further implementation details can be found in \Cref{adx:details:hparams}.

\paragraph{Evaluation Setup.}
We evaluate classifiers on two aspects: task performance and uncertainty awareness.
For \textit{text classification}, we report Macro-F1 and accuracy. 
For \textit{uncertainty quantification} we report Brier score (BS; \citealp{brier1950verification}), Expected Calibration Error (ECE; \citealp{naeini2015obtaining}), Static Calibration Error (SCE; \citealp{nixon2019measuring}), Negative log likelihood (NLL), coverage (Cov\%) and entropy ($H$). 
See \Cref{adx:details:calibration} for definitions.

\begin{figure}[th]
    \centering

    \subfloat[\label{fig:highlighted-main-results:entropy}Entropy]{
        \includegraphics[height=0.37\columnwidth]{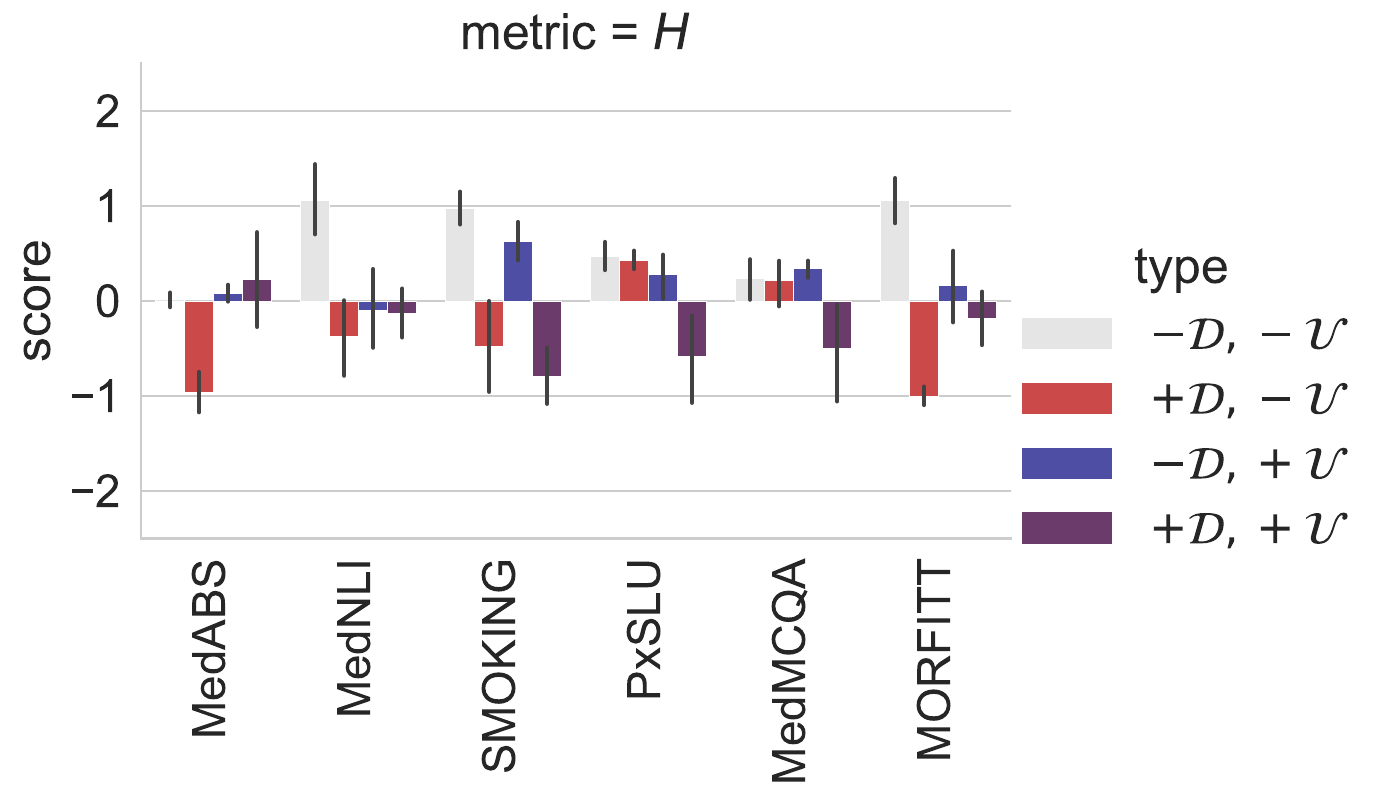}
    }

    \subfloat[\label{fig:highlighted-main-results:classification}Classification metrics]{
        \includegraphics[height=0.37\columnwidth]{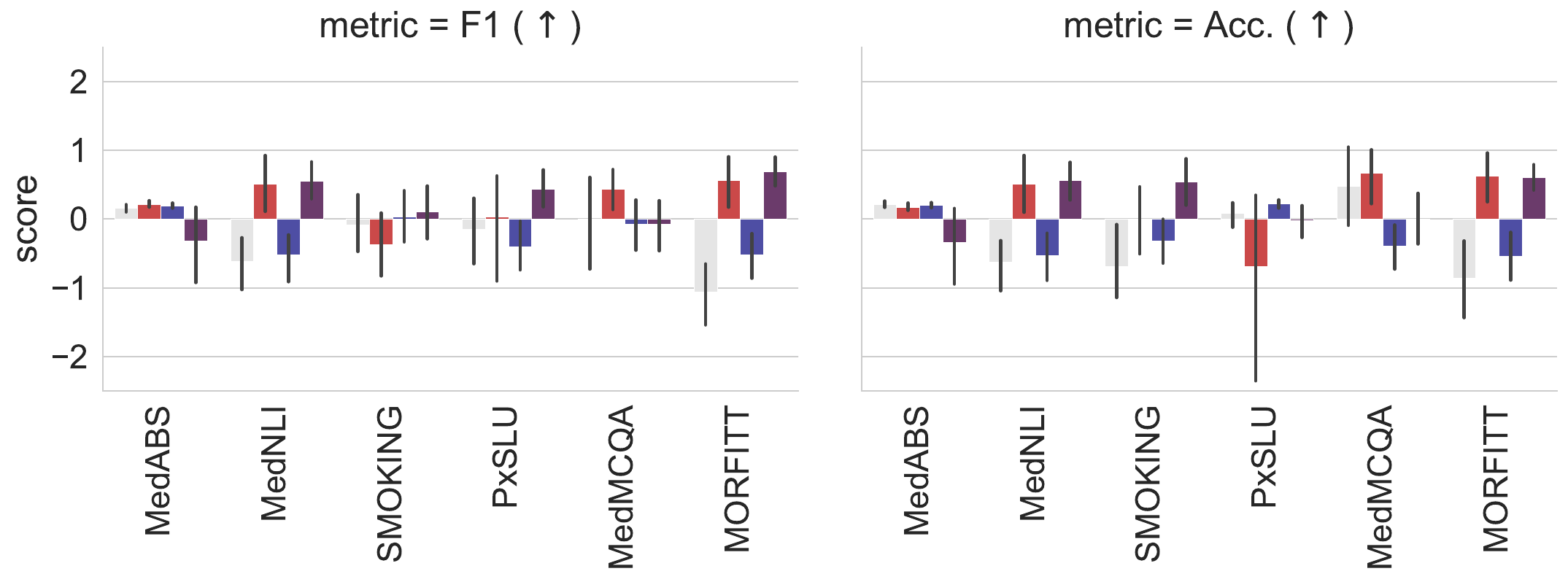}
    }

    \subfloat[\label{fig:highlighted-main-results:calibration}Calibration metrics]{
        \includegraphics[height=0.37\columnwidth]{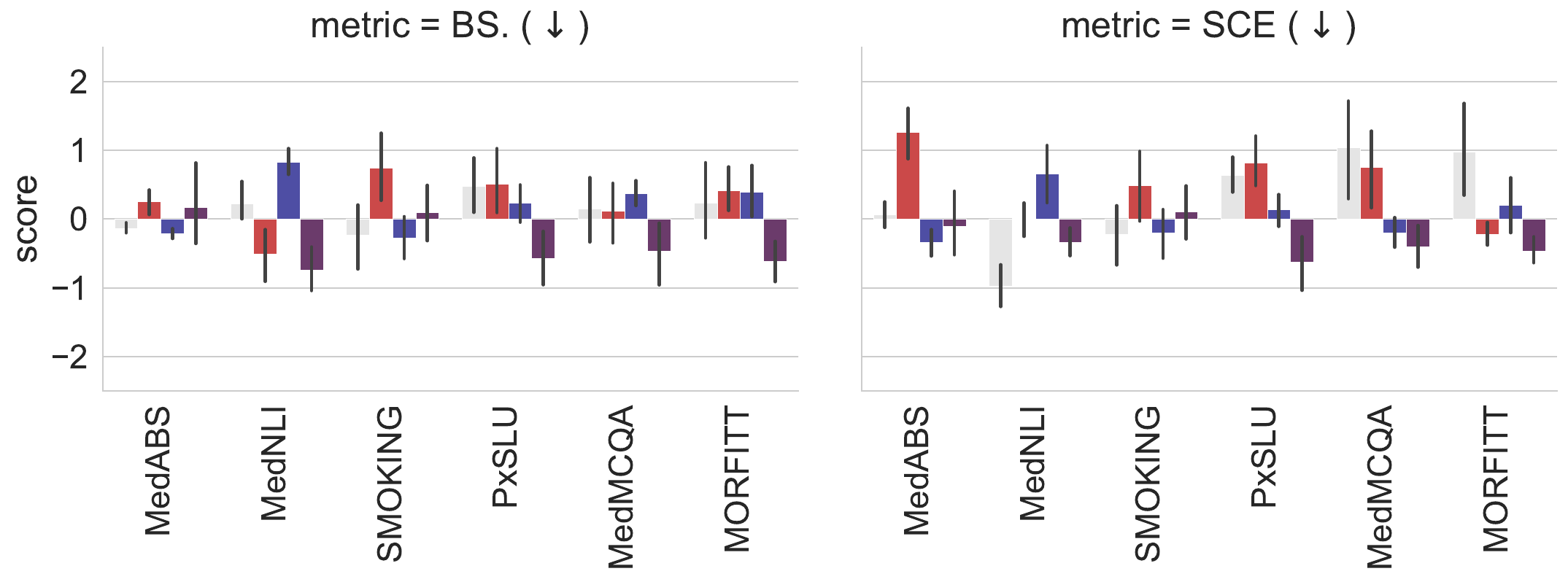}
    }

    \caption{
    Performances for empirically best models (selected metrics),
    $z$-normalized per dataset. See \Cref{tab:final-all-seeds} in \Cref{adx:sup results} for full non-normalized results.}
    \label{fig:highlighted-main-results}
\end{figure}

\section{Results}
\begin{figure*}[!h]
    \centering
    \subfloat[F1]{
        \includegraphics[width=0.24\linewidth]{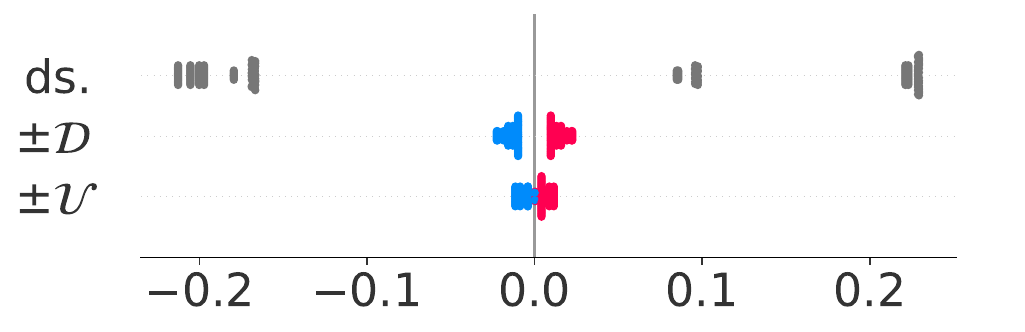}
    }
    \subfloat[Acc.]{
        \includegraphics[width=0.24\linewidth]{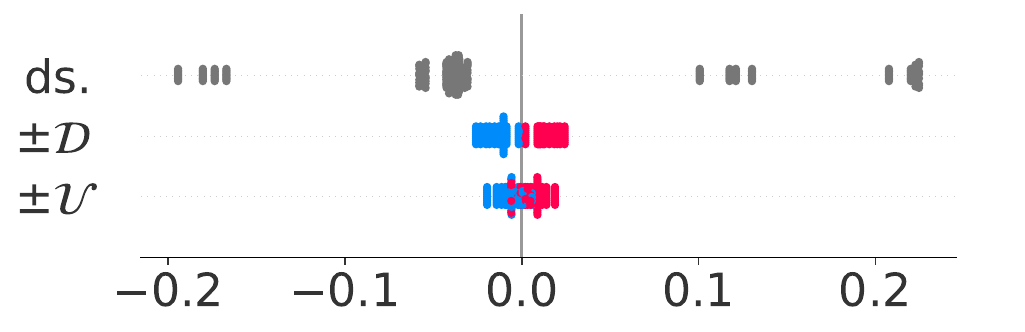}
    }
    \subfloat[NLL]{
        \includegraphics[width=0.24\linewidth]{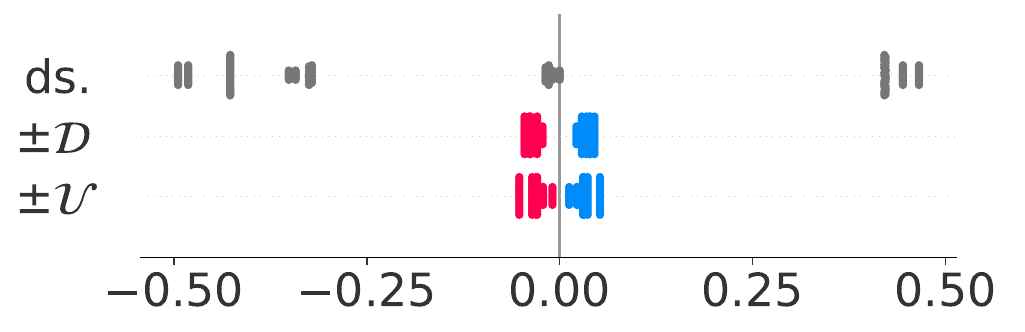}
    }
    \subfloat[$H$]{
        \includegraphics[width=0.24\linewidth]{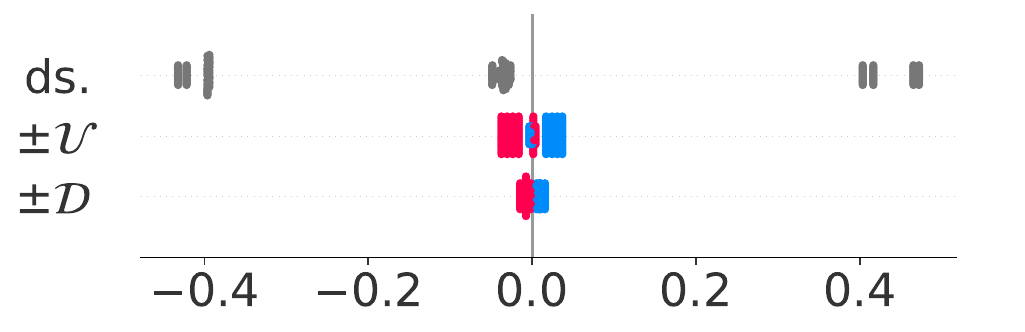}
    }

    \subfloat[Brier score]{
        \includegraphics[width=0.24\linewidth]{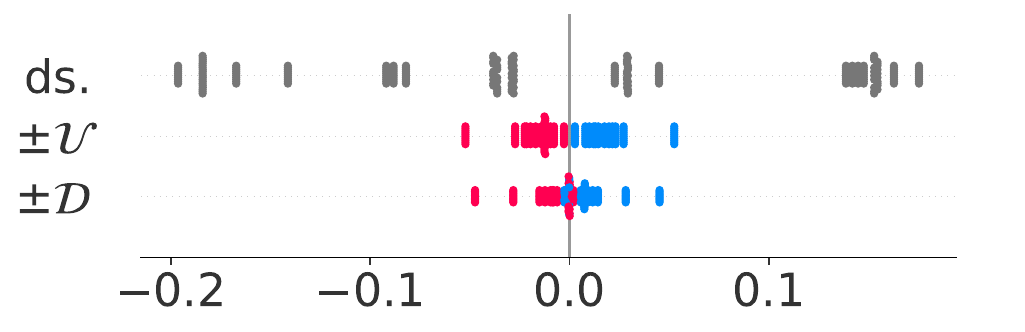}
    }
    \subfloat[SCE]{
        \includegraphics[width=0.24\linewidth]{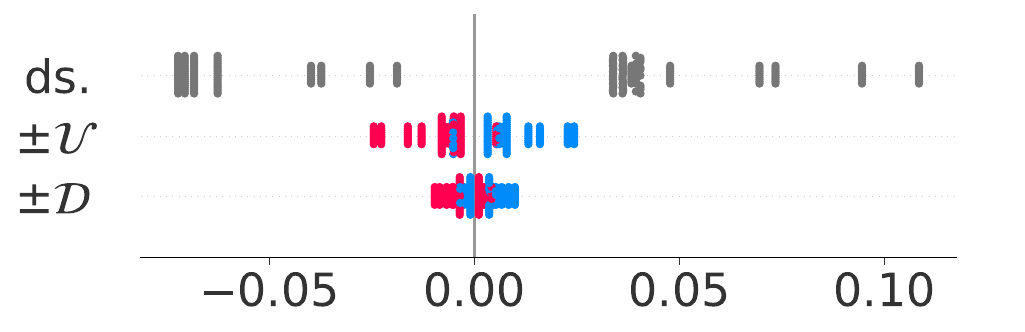}
    }
    \subfloat[ECE]{
        \includegraphics[width=0.24\linewidth]{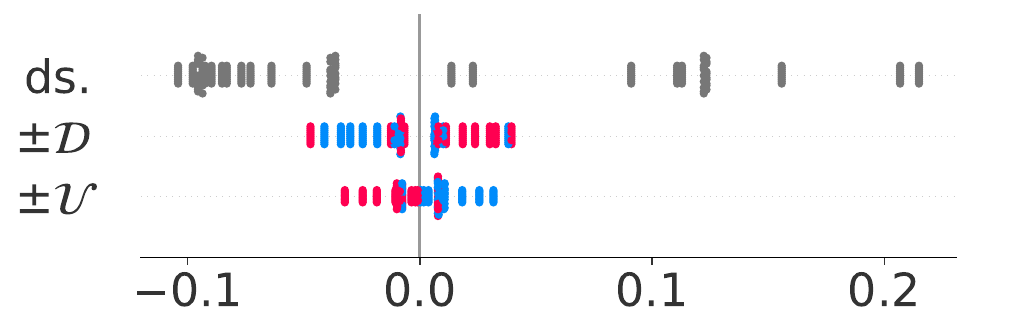}
    }
    \subfloat[Cov \%]{
        \includegraphics[width=0.24\linewidth]{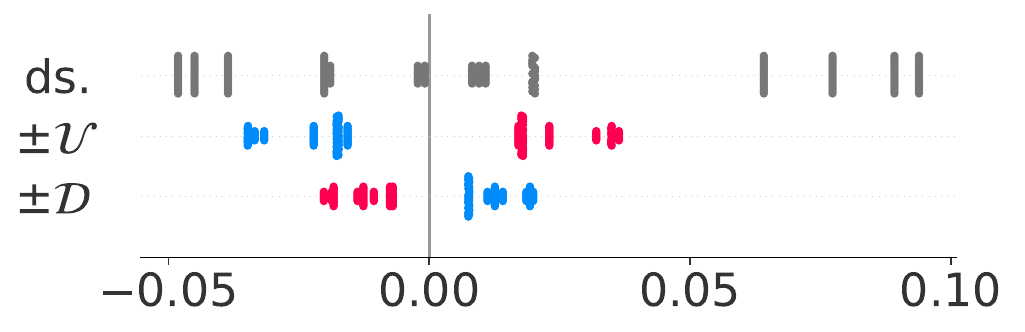}
    }

    \caption{SHAP attributions. Variables are ordered by mean absolute SHAPs. In blue, weight assigned when the variable is negative; in red, when it is positive. `ds.' denotes a categorical variable tracking the dataset.}
    \label{fig:some-shaps}
\end{figure*}

\paragraph{Performance.}
All results are listed in \Cref{tab:final-all-seeds} in \Cref{adx:sup results}, we highlight some key metrics in \Cref{fig:highlighted-main-results}.
\textcolor{black}{
Insofar as classification metrics go, $+\mathcal{D}$ configurations outperform  $-\mathcal{D}$ ones.
}
\textcolor{black}{More generally, a}s all scores are highly dependent on the exact dataset considered, we first de-trend them by $z$-normalizing results on a per-dataset basis \textcolor{black}{to simplify analysis}.
We find $+\mathcal{D} +\mathcal{U}$ classifiers to be strong contenders, although they are often outperformed---especially by $+\mathcal{D} -\mathcal{U}$ models on classification metrics (\Cref{fig:highlighted-main-results:classification}) and by $-\mathcal{D} +\mathcal{U}$ models on calibration metrics (\Cref{fig:highlighted-main-results:calibration}).
As for entropy, we find both $+\mathcal{D} -\mathcal{U}$ and  $+\mathcal{D} +\mathcal{U}$ to lead to lower scores.
Trends are consistent across languages.

    
    

\paragraph{Relative importance.} 
To interpret results in \Cref{fig:highlighted-main-results} more rigorously, we rely on SHAP \cite{NIPS2017_7062}.
SHAP is an algorithm to compute heuristics  for Shapley values \citep{shapley:book1952}, viz. a game theoretical additive and fair distribution of a given variable to be explained across predetermined factors of interest.
Here, we analyze the scores obtained by individual classifiers on all 8 metrics, and try to attribute their values ($z$-normalized per dataset) to domain specificity ($\pm\mathcal{D}$), uncertainty awareness ($\pm\mathcal{U}$) and the dataset one observation corresponds to (ds.). 

Results are displayed in \Cref{fig:some-shaps}; specific points correspond to weights assigned to one of the factors for one of the datapoints, factors are sorted from most to least impactful from top to bottom.
We can see that which of domain specificity and uncertainty awareness has the strongest impact depends strictly on the metrics: 
Cases where  $\pm\mathcal{D}$ is assigned on average a greater absolute weight than $\pm\mathcal{U}$ account for exactly half of the metrics we study.
Another import trend is that effects tied to $+\mathcal{D}$ are also often attested for $+\mathcal{U}$: if domain specificity is useful, then uncertainty awareness is as well.\footnote{ 
    There are two notable exceptions: ECE and coverage, where we find $+\mathcal{D}$ to be \emph{detrimental}.
    Variation across seeds might explain the discrepancy with \Cref{tab:final-all-seeds}.
}
Lastly, weights assigned to both $\pm\mathcal{D}$ and $\pm\mathcal{U}$ are considerably smaller than those assigned to datasets, showcasing that these trends are often overpowered by the specifics of the task at hand.

\begin{figure}[t]
    \centering
    \includegraphics[width=\columnwidth]{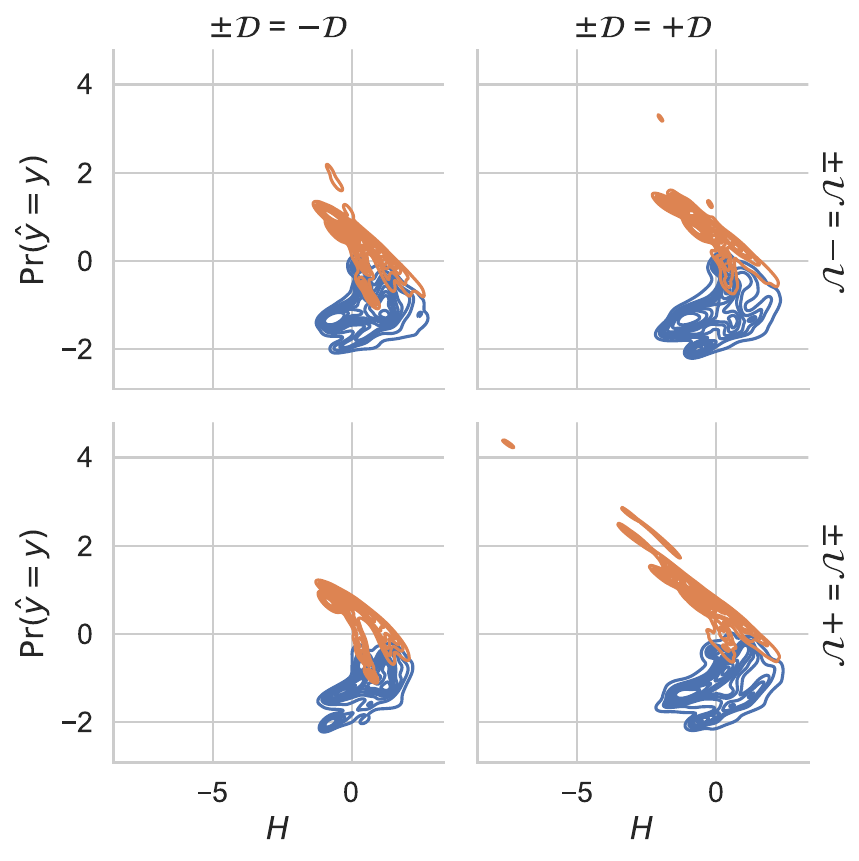}
    \caption{Entropy vs. probability mass assigned to the target ($z$-normalized per classifier). Orange: correct predictions; Blue: incorrect.}
    \label{fig:entropy-summary}
\end{figure}

\paragraph{Entropy.}
A desideratum we laid out above is to have large entropy scores when the model is incorrect.
Focusing on entropy, we display how it compares to the probability mass assigned to the target in \Cref{fig:entropy-summary}.
In detail, we retrieve all predictions for every datapoint across all classifiers and then $z$-normalize entropy scores and probability assigned the target class.\footnote{
    When plotting entropy against probability mass assigned to the target class, we can keep in mind some useful points of reference. 
    A perfect classifier that is always confidently correct should display a high probability mass and a low entropy (i.e., top left of our plot); what we hope to avoid is a confidently incorrect classifier (bottom left). 
    As entropy and probability are statistically related, it is impossible to observe a high probability mass and a high entropy (top right). Lastly, assuming the classifier outputs continuous scores, this statistical dependency also dictates that probability mass and entropy be inversely correlated for correct predictions.  
    }
{\color{black} We can see that incorrect predictions do result in more spread out entropy scores.
Moreover, w}e can notice some tentative differences between the four types of classifiers of our study: 
Correct predictions from
$+\mathcal{D} +\mathcal{U}$ models seem to lead to an especially tight correlation between entropy and probability mass. 

\begin{table}[t]
    \centering
    \resizebox{\columnwidth}{!}{\sisetup{
table-format = 3.1,
round-mode=places, 
round-precision=1,
detect-weight=true}
\def\Uline#1{#1\llap{\uline{\phantom{#1}}}}
\begin{tabular}{>{\bf}l@{{~}} S@{{~~}}S@{{~~}}S@{{~~}}S@{{\qquad}} S@{{~~}}S@{{~~}}S@{{~~}}S}
\toprule
 & \multicolumn{4}{c}{effect size $f$} & \multicolumn{4}{c}{Spearman's $\rho$} \\
 & {\rotatebox[origin=c]{90}{$-\mathcal{D}~-\mathcal{U}$}} & {\rotatebox[origin=c]{90}{$+\mathcal{D}~-\mathcal{U}$}} & {\rotatebox[origin=c]{90}{$-\mathcal{D}~+\mathcal{U}$}} & {\rotatebox[origin=c]{90}{$+\mathcal{D}~+\mathcal{U}$~}}
 & {\rotatebox[origin=c]{90}{$-\mathcal{D}~-\mathcal{U}$}}& {\rotatebox[origin=c]{90}{$+\mathcal{D}~-\mathcal{U}$}} & {\rotatebox[origin=c]{90}{$-\mathcal{D}~+\mathcal{U}$}} & {\rotatebox[origin=c]{90}{$+\mathcal{D}~+\mathcal{U}$}}  \\
\midrule
MedABS   &      62.4587 & \Uline{64.8} &      62.3861 &  \bf 67.3204 & \Uline{-48.0} &      -47.9223 &      -44.6260 &  \bf -53.5326 \\
MedNLI   &      73.2472 &      73.2104 & \Uline{74.0} &  \bf 76.9906 &      -73.1618 & \Uline{-77.4} &      -76.0566 &  \bf -83.2975 \\
SMOKING  &  \bf 75.8140 &      71.5580 &      74.1591 & \Uline{74.8} &  \bf -56.4635 &      -37.9853 &      -50.0037 & \Uline{-56.0} \\
PxSLU &      65.4190 &  \bf 87.2314 &      65.0581 & \Uline{85.8} &      -85.4051 &      -69.0819 & \Uline{-87.3} &  \bf -96.1698 \\
MedMCQA  &      65.5612 &      63.8467 & \Uline{66.6} &  \bf 68.1908 &  \bf -82.2927 & \Uline{-82.2} &      -60.7625 &      -62.5836 \\
MORFITT  & \Uline{65.6} &  \bf 66.1075 &      65.0232 &      64.7896 & \Uline{-54.6} &  \bf -55.0504 &      -50.7645 &      -50.9648 \\
\bottomrule
\end{tabular}}
    \caption{Statistical tests on entropy measurements, with \textbf{best} and \underline{second best} highlighted.}
    \label{tab:ent-stats}
\end{table}

However, establishing whether this difference is significant requires further testing.
We therefore measure whether incorrect predictions lead to higher entropy in two ways: (i) using Mann--Whitney U-tests, from which we derive a common language effect size $f$ (as the entropy of incorrect predictions should be higher);\footnote{All U-tests suggest entropy for incorrect predictions is significantly higher ($p < 10^{-10}$).} 
and (ii), by computing Spearman correlation coefficients between the entropy and the mass assigned to the target class (as entropy should degrade with correctness).
Corresponding results are listed in \Cref{tab:ent-stats}: 
Across most of the datasets we study, the top or second most coherent distributions  we observe are for domain-specific and uncertainty-aware models.
However, we also observe that actual performances are highly sensitive to the exact classification task at hand.







\section{Discussion \& Conclusion}
We can now answer our initial research questions.

\textbf{RQ1}: \textit{Are the benefits of uncertainty-awareness and domain-specificity orthogonal?} 
We have seen in \Cref{tab:ent-stats} that in most cases, using a classifier that was both domain-specific and uncertainty-aware led to the optimal distribution shape, with entropy more gracefully increasing with incorrectness.

\textbf{RQ2}: \textit{Should medical practitioners prioritize domain-specificity or uncertainty-awareness?}
SHAP attributions in \Cref{fig:some-shaps} strongly suggest that the evaluation metric dictates the strategy to follow. 
As one would expect, accuracy is better captured with domain-specific models, whereas uncertainty-aware models tend to be better calibrated.

We also found
significant evidence throughout our experiments that the exact classification task at hand weighs in much more strongly than the design of the classifier. 
This extraneous factor necessarily complicates the relationship between domain-specificity and uncertainty-awareness: 
In a handful of cases in \Cref{fig:highlighted-main-results}, we observe classifiers that are neither uncertainty-aware nor domain specific faring best among all the models we survey---and conversely domain-specific uncertainty-aware classifiers can also rank dead last.
This is also related to the often limited quantitative difference between best and worst models, which for instance can be as low as $\pm2.3\%$ for F1 on MEDABS (cf. \Cref{tab:final-all-seeds}).

Overall, our experiments suggest a very nuanced conclusion.
Domain-specificity and uncertainty-awareness do appear to shape classifiers' distributions and their entropy in distinct but compatible ways, but they have a lesser impact than the task itself.
Hence, while we can often combine uncertainty-awareness and domain-specificity, there are no out-of-the-box solutions, and optimal performances require careful application designs.

\section*{Acknowledgments}
We thank Sami Virpioja for his comments on an early version of this work.

This work is supported by the ICT 2023 project ``Uncertainty-aware neural language models'' funded by the Academy of Finland (grant agreement  \textnumero{}~345999).


\bibliography{custom}

\appendix
\begin{figure*}
    \centering
    \includegraphics[scale=0.48]{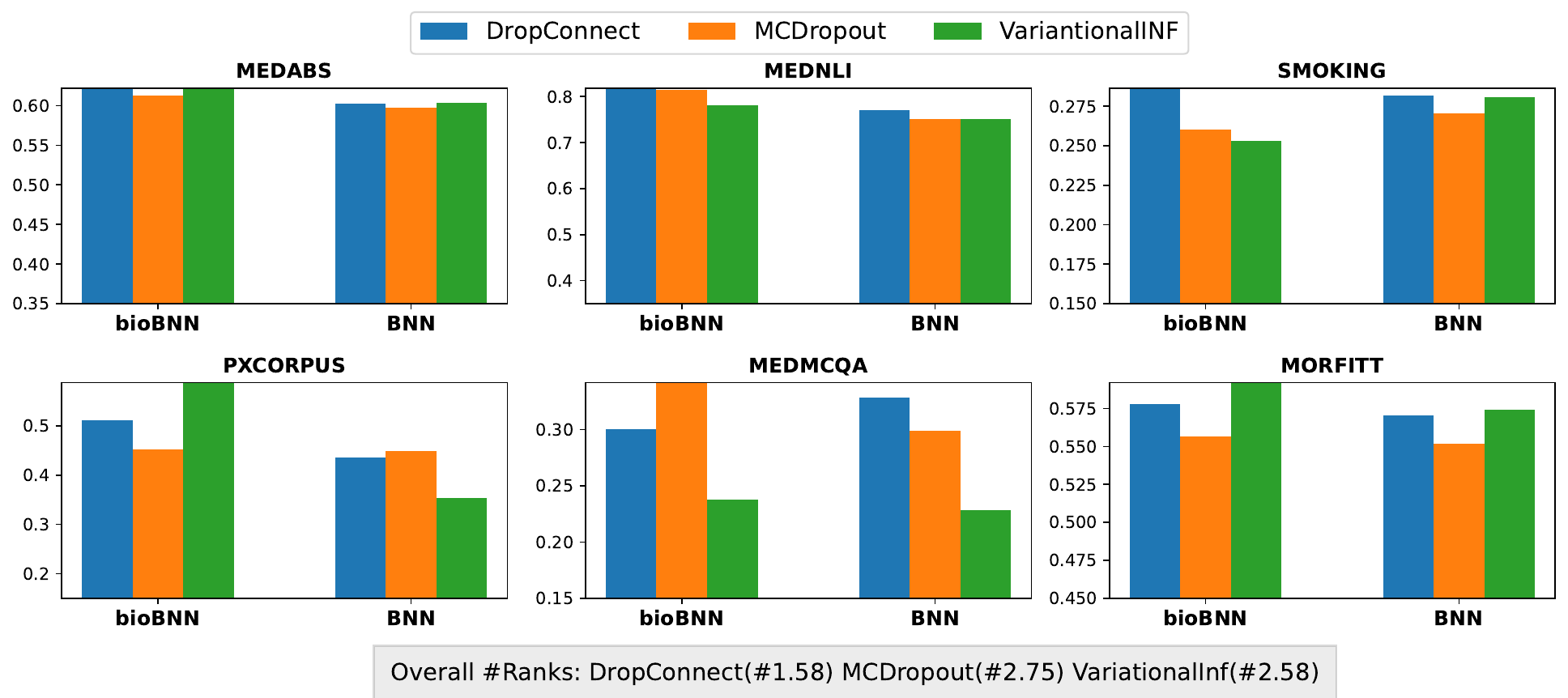}
    \caption{Comparison of various BNN models for different datasets on classification task based on Macro-F1 on validation set. }
    \label{fig:compare-rank-for-selection}
\end{figure*}
\section{Experimental details}
\label{adx:details}

\subsection{Supplementary Bayesian models}
\label{adx:details:models}

We include the details for two more Bayesian models: MC-dropout and variational inference. Note that for all the Bayesian models we sample K=3 predictions at inference and use the mean prediction. 

\paragraph{MCDropout (MCD)} This model is based on a PLM encoder, similar to the main study models. 
The difference in this case is that Stochastic Dropout  is applied over the classification layer.
MCD \cite{gal2016dropout} proposes to extend the usage of Dropout but at inference time enabling it to sample a multiple $K$
models, to make $K$ predictions. The final prediction in the case of classification model can denoted as
\[
\hat{y} = K^{-1}\sum_{k=1}^K f_i(x)
\]. 

\paragraph{Variational inference (VI)} This model is based on a PLM encoder, similar to the main study models, with variational inference dense layer as the classification layer. 
We use the Bayes by BackProp \cite{blundell2015weight} for the VI Dense layer. It approximates the distribution of each weight with a Gaussian distribution with parameter $\mathcal{N}(\mu, \rho)$. The weights are approximated with Monte Carlo gradient. Finally, the predictions are computed using the predictive posterior distribution, by sampling K weight instances and making one forward pass per set of weights same as MCD.


\subsection{Implementation details}
\label{adx:details:hparams}

We use \href{https://github.com/mvaldenegro/keras-uncertainty}{\tt keras-uncertainty} models for implementing our BNN model backbones.

\begin{table}[]
    \centering
    \resizebox{\linewidth}{!}{
    \begin{tabular}{rr cc c cc c cc c cc c cc c cc}
    \toprule
        \multicolumn{2}{c}{Models} & \multicolumn{2}{c}{MedABS} & & \multicolumn{2}{c}{MedNLI} & & \multicolumn{2}{c}{SMOKING} & & \multicolumn{2}{c}{PxSLU} & & \multicolumn{2}{c}{MedMCQA} & & \multicolumn{2}{c}{MORFITT} \\
        \cmidrule{3-4} \cmidrule{6-7} \cmidrule{9-10} \cmidrule{12-13} \cmidrule{15-16} \cmidrule{18-19}
         &&lr & E & &lr & E& &lr & E& &lr & E& &lr & E& &lr & E \\
         \midrule
         $-\mathcal{D}$ &DNN  & 1e-5 &4 && 5e-6 &12 &&1e-4 & 15 &&5e-6 &15 && 5e-6 &14 && 5e-5 &15\\
         $-\mathcal{D}$ &DC   & 5e-6 &7 && 1e-5 &11	&&1e-5 & 15 &&5e-6 &13 && 5e-6 &15 && 5e-5 &13\\
         $-\mathcal{D}$ &MCD  & 5e-5 &5  && 5e-6 &15  &&5e-5 & 15 &&1e-5 &14 && 5e-6 &11 && 5e-5 &10\\
         $-\mathcal{D}$ &VI   & 5e-6 &7 && 1e-5 &14 &&5e-6 & 13 &&5e-6 &14 && 1e-6 &15 && 5e-5 &13\\
         \midrule
         $+\mathcal{D}$ &DNN  & 1e-5 & 4&& 1e-5& 14 &&5e-5 & 15 &&1e-5 &15 && 1e-5 &10 && 5e-5 &15\\
         $+\mathcal{D}$ &DC   & 5e-5 & 3&& 1e-5 &13 &&1e-4 & 13 &&1e-5 &15 && 5e-6 &15 && 5e-5 &13\\
         $+\mathcal{D}$ &MCD  & 5e-5 & 3&& 5e-5 &12 &&5e-5 & 10 &&1e-5 &14 && 1e-5 &15 && 5e-5 &13\\
         $+\mathcal{D}$ &VI   & 1e-5 & 5&& 5e-6 &13 &&5e-5 & 14	&&1e-5 &14 && 1e-6 &15 && 5e-5 &5\\
         \bottomrule
         
    \end{tabular}
    }
    \caption{Best hyparameter for each model configuration and dataset pair. We denote both English and French domain-specific PLMs with $+\mathcal{D}$. The models DC, MCD, VI are from the $+\mathcal{U}$ set.} 
    \label{tab:BEST-HYP}
\end{table}

\paragraph{Hyperparameter Setting}
In all cases, we fine-tune the PLM backbone for all the downstream task with a maximum sequence length of 512 and a batch size of 50 sentences. 
We perform a grid hyper-parameter search for \texttt{epochs}= \{3,4,5, ..., 15\} and \texttt{lr}= \{1e-7, 5e-6, 1e-6, 5e-5, 1e-5, 5e-4, 1e-4\}. We replicate training with 3 seeds for each hyperparameter configuration, select the optimal configuration for validation F1, and replicate training with 7 more seeds for these optimal configurations, so as to obtain 10 models per dataset, PLM and architecture. We also select the main BNN model of the study by selecting the system yielding the highest average rank across all six datasets, as displayed in \Cref{fig:compare-rank-for-selection}.

We train all  models with binary cross entropy loss and \texttt{Adam} optimizer with $\epsilon=10^{-8}$ and $\beta=(0.9, 0.999)$. For all BNN models, we obtain 3 sets of predictions after training the models to calculate the mean class probabilities.
Corresponding optimal hyperparameters are listed in \cref{tab:BEST-HYP}.



\subsection{Calibration metrics definition}
\label{adx:details:calibration}

In what follows, $N$ denotes the number of samples in test set, $C$ denotes the number of classes. Lower score for Brier score, ECE, SCE, NLL and Entropy metrics; and higher score for coverage, are indicative of better uncertainty aware model.  

\paragraph{Brier score.} \citet{brier1950verification} proposed BS which computes the mean square difference between the true classes and the predicted probabilities. 
$$\text{BS} = \frac{1}{N}\sum^{N}_{i=1}\sum^{C}_{c=1} (y^{(i)}_c  - \hat{y}^{(i)}_c)^2$$





\paragraph{Expected Calibration Error.} \citet{naeini2015obtaining} provides weighted average of the difference between accuracy and confidence across  $B$ bins. 
$$\text{ECE} = \sum^{B}_{b=1} \frac{n_b}{N} |\text{acc}(b)-\text{conf}(b)|$$
where acc$(b)$ and conf$(b)$ are the average accuracy and confidence of predictions in bin $b$, respectively. We set $B = 15$ in our experiments.

\paragraph{Static Calibration Error.} \citet{nixon2019measuring} proposed an extension of ECE to multi-class problems to overcome its limitation of dependence of the number of bins. 
$$\text{SCE} = \sum^{C}_{c=1} \sum^{B}_{b=1} \frac{n_b}{NC} |\text{acc}(b)-\text{conf}(b)|$$
We set $B = 15$ in our experiments.

\paragraph{Negative Log Likelihood.} serves as the primary approach for optimizing neural networks in classification tasks. Interestingly, this loss function can also double as an effective metric for assessing uncertainty. 
$$\text{NLL} = - \sum_{i=1}^{N} y_i \log(\hat{y}_i)$$

\paragraph{Coverage Percentage.} The normalized form of number of times the
correct class in indeed contain within the prediction
set.

\paragraph{Shannon Entropy.} quantifies the expected uncertainty inherent in the possible outcomes of a discrete random variable.
$$H = - \sum_{i=1}^{N} p_i \log(p_i)$$
\subsection{Dataset}
\label{adx:sup datasets}
We provided supplementary details about each dataset we used in Table \ref{tab:datasample}.
\begin{table*}[]
    \centering
    \resizebox{\textwidth}{!}{%
    \begin{tabular}{ccp{8cm}p{4cm}p{4cm}}
        \hline
        \textbf{Dataset}  && \textbf{Sample}  & \textbf{Classes}&\textbf{Label Distribution}\\
        \hline
        MedABS \cite{10.1145/3582768.3582795} &  &\small{\{\texttt{\textcolor{blue}{text}}: "\textit{Catheterization of coronary artery bypass graft from the descending aorta. The increasing frequency of reoperation for coronary artery disease has led to the use of a variety of grafts. This report describes the catheter technique for selective opacification of a saphenous vein graft from the descending thoracic aorta to the posterior coronary circulation. }", \texttt{\textcolor{blue}{label}}: "Cardiovascular diseases" \}} &\{\small{'Neoplasms', 'Digestive system', 'Nervous system', 'Cardiovascular', 'General pathological'} \}& [1925 913 1149 1804 2871]\\
        MedNLI \cite{romanov2018lessons} & &\small{\{\texttt{\textcolor{blue}{text}}: "\textit{No history of blood clots or DVTs, has never had chest pain prior to one week ago. [SEP] Patient has angina}",  \texttt{\textcolor{blue}{label}}: "entailment"\}
}&\{\small{"entailment", "contradict", "neutral"} \}& [3744 3744 3744]\\
        SMOKING \cite{uzuner2008identifying}&&\small{\{\texttt{\textcolor{blue}{text}}: "\textit{071962960 bh 4236518 417454 12/10/2001 12:00:00 am discharge summary unsigned dis report status : unsigned discharge summary name : sterpsap , ny unit number : 582-96-88 admission date : 12/10/2001 discharge date : 12/19/2001 principal diagnosis : prosthetic aortic valve dysfunction associated diagnoses : aortic valve insufficiency bacterial endocarditis , active principal procedure : urgent re-do aortic valve replacement and correction of left ventricular to aortic discontinuity . ( 12/13/2001 ) other procedures : aortic root aortogram ( 12/12/2001 ) cardiac ultrasound ( 12/13/2001 ) insertion dual chamber pacemaker ( 12/15/2001 ) picc line placement ( 12/18/2001 ) history and reason for hospitalization : mr. sterpsap ...}",  \texttt{\textcolor{blue}{label}}: "CURRENT SMOKER"\}}& \{\small{'CURRENT SMOKER', 'NON-SMOKER', 'PAST SMOKER', 'SMOKER', 'UNKNOWN'} \}&[ 27 49 24 8 190] \\
        
        MEDMCQA \cite{labrak2023frenchmedmcqa}&&\small{\{\texttt{\textcolor{blue}{text}}: "\textit{ans la liste suivante, quels sont les antibiotiques utilisables pour traiter une salmonellose chez un adulté?}", \texttt{\textcolor{blue}{label}}: 2\}} &\{\small{1,2,3,4,5}\}& [595 528 718 296 34] \\
        
        MORFITT \cite{labrak:hal-04131591}&&\small{\{\texttt{\textcolor{blue}{text}}: "\textit{La survenue de complications postopératoires représente un cauchemar (bien réel), tant pour le patient que pour son chirurgien. Dès lors, quoi de plus fantasmagorique que d’administrer une « potion magique » au patient avant l’intervention pour éliminer ce risque ? Le but de cet article est de résumer l’état des connaissances actuelles concernant les bénéfices potentiels, liés à l’administration d’immunonutrition aux patients traités pour cancer urologique.....}", \texttt{\textcolor{blue}{original\_label}}: [ "Immunologie","Chirurgie",],  \texttt{\textcolor{blue}{label}}: "Immunologie"\}}&\{\small{'Vétérinaire', 'Étiologie', 'Psychologie', 'Chirurgie', 'Génétique', 
        'Physiologie', 'Pharmacologie', 'Microbiologie', 'Immunologie', 'Chimie',                        'Virologie', 'Parasitologie'} \} &[ 82 261 32 122 40 17 152 39 242 185 104 238] \\
        
        PxSLU \cite{Kocabiyikoglu2022}& & \small{\{\texttt{\textcolor{blue}{text}}: "\textit{antacapone 200 milligrammes 2 comprimés le matin 1 comprimé à midi 2 comprimé le soir traitement pour une durée totale de 4 semaines}", \texttt{\textcolor{blue}{label}}: "medical\_prescription"\}}  &\{\small{"medical\_prescription", "negate","replace", "none"} \}& [1276 15 82 13]\\
         \hline
    \end{tabular}}
    \caption{Sample data from each Dataset}
    \label{tab:datasample}
\end{table*}




\begin{table*}[!th]
    \centering
    \resizebox{\textwidth}{!}{%
    \begin{tabular}{r cc ccccccccc}
    \toprule
       & \multicolumn{2}{c}{\multirow{2}{*}{Model}} &  \multicolumn{2}{c}{Classification}& &  \multicolumn{6}{c}{Uncertainty}\\
        \cmidrule{4-5} \cmidrule{7-12}
            &&& Macro-F1($\uparrow$) & Accuracy($\uparrow$)& & BS($\downarrow$) & ECE($\downarrow$) & SCE($\downarrow$)  & NLL($\downarrow$) & Cov\%($\uparrow$) & Entropy($\downarrow$) \\
    \midrule
    
         \parbox[t]{2mm}{\multirow{8}{*}{\rotatebox[origin=c]{90}{MedABS \resizebox{2ex}{!}{\worldflag{US}}}}} & 
          $-\mathcal{D}$  & DNN&60.3633$\pm$\small{0.003}&60.9765$\pm$\small{0.002}&&0.5535$\pm$\small{0.008}&0.1387$\pm$\small{0.016}&0.0683$\pm$\small{0.004}&1.3261$\pm$\small{0.001}&0.8976$\pm$\small{0.013}&1.5579$\pm$\small{0.002}\\
& $-\mathcal{D}$&DC&60.9855$\pm$\small{0.004}&61.1842$\pm$\small{0.003}&&0.5518$\pm$\small{0.002}&\underline{0.1342}$\pm$\small{0.007}&0.0674$\pm$\small{0.003}&1.3192$\pm$\small{0.002}&0.9611$\pm$\small{0.003}&1.5556$\pm$\small{0.001}\\
& $-\mathcal{D}$&MCD&60.6979$\pm$\small{0.004}&60.0993$\pm$\small{0.006}&&0.5691$\pm$\small{0.015}&0.1503$\pm$\small{0.014}&0.0688$\pm$\small{0.01}&1.3235$\pm$\small{0.008}&0.9401$\pm$\small{0.013}&1.5542$\pm$\small{0.002}\\
& $-\mathcal{D}$&VI&60.8725$\pm$\small{0.001}&61.1611$\pm$\small{0.001}&&0.5531$\pm$\small{0.006}&0.1394$\pm$\small{0.004}&0.0695$\pm$\small{0.003}&1.3164$\pm$\small{0.003}&0.958$\pm$\small{0.001}&1.5541$\pm$\small{0.001}\\
\cmidrule{2-12}
&$+\mathcal{D}$  &DNN&60.8077$\pm$\small{0.013}&61.3343$\pm$\small{0.01}&&0.5499$\pm$\small{0.014}&0.1448$\pm$\small{0.005}&0.0695$\pm$\small{0.001}&1.3201$\pm$\small{0.014}&0.9193$\pm$\small{0.005}&1.5561$\pm$\small{0.003}\\
&$+\mathcal{D}$  &DC&\underline{62.5642}$\pm$\small{0.009}&62.1018$\pm$\small{0.01}&&\underline{0.5243}$\pm$\small{0.015}&0.1381$\pm$\small{0.016}&\underline{0.0624}$\pm$\small{0.007}&\underline{1.2962}$\pm$\small{0.007}&\underline{0.9597}$\pm$\small{0.008}&\underline{1.5523}$\pm$\small{0.002}\\
&$+\mathcal{D}$  &MCD&62.2038$\pm$\small{0.022}&\underline{62.1307}$\pm$\small{0.022}&&0.5226$\pm$\small{0.031}&\textbf{0.1238}$\pm$\small{0.031}&\textbf{0.0593}$\pm$\small{0.015}&1.3056$\pm$\small{0.013}&\textbf{0.9666}$\pm$\small{0.01}&1.5562$\pm$\small{0.002}\\
&$+\mathcal{D}$  &VI&\textbf{63.1893}$\pm$\small{0.004}&\textbf{63.1694}$\pm$\small{0.003}&&\textbf{0.5234}$\pm$\small{0.009}&0.1464$\pm$\small{0.01}&0.0653$\pm$\small{0.003}&\textbf{1.288}$\pm$\small{0.006}&0.9603$\pm$\small{0.005}&\textbf{1.5491}$\pm$\small{0.002}\\
\midrule
\parbox[t]{2mm}{\multirow{8}{*}{\rotatebox[origin=c]{90}{MedNLI \resizebox{2ex}{!}{\worldflag{US}}}}}
& $-\mathcal{D}$&DNN&73.8951$\pm$\small{0.013}&73.8397$\pm$\small{0.015}&&0.3976$\pm$\small{0.006}&0.1278$\pm$\small{0.02}&0.0846$\pm$\small{0.012}&0.8177$\pm$\small{0.015}&\underline{0.9119}$\pm$\small{0.008}&1.0156$\pm$\small{0.008}\\
& $-\mathcal{D}$&DC&74.8161$\pm$\small{0.019}&74.8711$\pm$\small{0.018}&&0.4242$\pm$\small{0.021}&0.185$\pm$\small{0.007}&0.1259$\pm$\small{0.005}&0.7945$\pm$\small{0.014}&0.8509$\pm$\small{0.007}&0.9941$\pm$\small{0.002}\\
& $-\mathcal{D}$&MCD&72.8896$\pm$\small{0.03}&73.0192$\pm$\small{0.03}&&0.4163$\pm$\small{0.009}&\underline{0.1214}$\pm$\small{0.049}&\underline{0.0865}$\pm$\small{0.03}&0.8298$\pm$\small{0.037}&0.9109$\pm$\small{0.04}&1.0171$\pm$\small{0.02}\\
& $-\mathcal{D}$&VI&73.0816$\pm$\small{0.022}&73.1364$\pm$\small{0.022}&&0.4426$\pm$\small{0.016}&0.185$\pm$\small{0.023}&0.1265$\pm$\small{0.015}&0.8109$\pm$\small{0.022}&0.857$\pm$\small{0.035}&0.9983$\pm$\small{0.011}\\
\cmidrule{2-12}
&$+\mathcal{D}$  &DNN&77.172$\pm$\small{0.041}&77.2386$\pm$\small{0.039}&&0.3783$\pm$\small{0.05}&0.1579$\pm$\small{0.009}&0.107$\pm$\small{0.007}&0.7736$\pm$\small{0.039}&0.857$\pm$\small{0.015}&0.9952$\pm$\small{0.008}\\
&$+\mathcal{D}$  &DC&\underline{79.9945}$\pm$\small{0.037}&\underline{80.0047}$\pm$\small{0.037}&&\textbf{0.3375}$\pm$\small{0.045}&0.1392$\pm$\small{0.005}&0.0956$\pm$\small{0.002}&\underline{0.7486}$\pm$\small{0.041}&0.8872$\pm$\small{0.011}&\underline{0.9924}$\pm$\small{0.011}\\
&$+\mathcal{D}$  &MCD&\textbf{80.1022}$\pm$\small{0.014}&\textbf{80.1688}$\pm$\small{0.014}&&\underline{0.3453}$\pm$\small{0.02}&0.1565$\pm$\small{0.009}&0.1065$\pm$\small{0.005}&\textbf{0.7437}$\pm$\small{0.012}&0.8654$\pm$\small{0.004}&\textbf{0.9872}$\pm$\small{0.001}\\
&$+\mathcal{D}$  &VI&77.0617$\pm$\small{0.043}&77.1027$\pm$\small{0.042}&&0.351$\pm$\small{0.046}&\textbf{0.1041}$\pm$\small{0.019}&\textbf{0.0773}$\pm$\small{0.01}&0.7851$\pm$\small{0.046}&\textbf{0.9293}$\pm$\small{0.025}&1.0101$\pm$\small{0.015}\\
\midrule
\parbox[t]{2mm}{\multirow{8}{*}{\rotatebox[origin=c]{90}{SMOKING \resizebox{2ex}{!}{\worldflag{US}}}}}
& $-\mathcal{D}$&DNN&\textbf{27.1141}$\pm$\small{0.041}&45.8333$\pm$\small{0.142}&&0.7724$\pm$\small{0.054}&0.2961$\pm$\small{0.057}&0.154$\pm$\small{0.012}&1.4298$\pm$\small{0.106}&0.7724$\pm$\small{0.163}&1.5536$\pm$\small{0.028}\\
& $-\mathcal{D}$&DC&25.7924$\pm$\small{0.041}&46.7949$\pm$\small{0.039}&&\textbf{0.6407}$\pm$\small{0.035}&\textbf{0.1625}$\pm$\small{0.043}&\textbf{0.1215}$\pm$\small{0.016}&1.4331$\pm$\small{0.035}&\underline{0.9455}$\pm$\small{0.031}&1.5791$\pm$\small{0.01}\\
& $-\mathcal{D}$&MCD&26.707$\pm$\small{0.058}&45.8333$\pm$\small{0.073}&&0.7609$\pm$\small{0.077}&0.2771$\pm$\small{0.048}&0.1507$\pm$\small{0.021}&1.4519$\pm$\small{0.045}&0.8942$\pm$\small{0.058}&1.5651$\pm$\small{0.003}\\
& $-\mathcal{D}$&VI&23.4485$\pm$\small{0.034}&32.0513$\pm$\small{0.043}&&0.7197$\pm$\small{0.053}&0.2171$\pm$\small{0.021}&0.15$\pm$\small{0.023}&1.5031$\pm$\small{0.038}&0.8974$\pm$\small{0.113}&1.5887$\pm$\small{0.004}\\
\cmidrule{2-12}
&$+\mathcal{D}$  &DNN&24.9822$\pm$\small{0.041}&\textbf{51.6026}$\pm$\small{0.071}&&\underline{0.6764}$\pm$\small{0.076}&0.2262$\pm$\small{0.013}&\underline{0.1334}$\pm$\small{0.031}&\underline{1.3928}$\pm$\small{0.068}&0.6571$\pm$\small{0.114}&1.5596$\pm$\small{0.011}\\
&$+\mathcal{D}$  &DC&\underline{27.0293}$\pm$\small{0.033}&47.1154$\pm$\small{0.075}&&0.841$\pm$\small{0.043}&0.3441$\pm$\small{0.053}&0.1738$\pm$\small{0.007}&1.4297$\pm$\small{0.06}&0.7276$\pm$\small{0.118}&\underline{1.5419}$\pm$\small{0.02}\\
&$+\mathcal{D}$  &MCD&25.0029$\pm$\small{0.051}&40.3846$\pm$\small{0.058}&&0.6777$\pm$\small{0.022}&\underline{0.206}$\pm$\small{0.019}&0.1401$\pm$\small{0.014}&1.482$\pm$\small{0.014}&\textbf{0.9487}$\pm$\small{0.04}&1.5895$\pm$\small{0.003}\\
&$+\mathcal{D}$  &VI&26.1167$\pm$\small{0.03}&\underline{50.3205}$\pm$\small{0.094}&&0.765$\pm$\small{0.175}&0.3201$\pm$\small{0.094}&0.1584$\pm$\small{0.045}&\textbf{1.3857}$\pm$\small{0.094}&0.75$\pm$\small{0.063}&\textbf{1.5397}$\pm$\small{0.003}\\
\midrule
\parbox[t]{2mm}{\multirow{8}{*}{\rotatebox[origin=c]{90}{PxSLU \resizebox{2ex}{!}{\worldflag{FR}}}}}
& $-\mathcal{D}$&DNN&32.2541$\pm$\small{0.075}&88.2452$\pm$\small{0.012}&&0.5743$\pm$\small{0.077}&0.4556$\pm$\small{0.094}&0.2955$\pm$\small{0.014}&1.2807$\pm$\small{0.05}&0.995$\pm$\small{0.004}&1.3821$\pm$\small{0.003}\\
& $-\mathcal{D}$&DC&34.1464$\pm$\small{0.026}&84.2989$\pm$\small{0.05}&&0.4599$\pm$\small{0.088}&0.3936$\pm$\small{0.047}&0.2354$\pm$\small{0.03}&1.2154$\pm$\small{0.062}&\textbf{1.0}$\pm$\small{0.0}&1.3768$\pm$\small{0.007}\\
& $-\mathcal{D}$&MCD&33.211$\pm$\small{0.067}&88.6902$\pm$\small{0.018}&&0.5232$\pm$\small{0.103}&0.4852$\pm$\small{0.079}&0.2615$\pm$\small{0.027}&1.2571$\pm$\small{0.062}&\textbf{1.0}$\pm$\small{0.0}&1.3806$\pm$\small{0.004}\\
&$-\mathcal{D}$&VI&25.9883$\pm$\small{0.041}&88.9169$\pm$\small{0.013}&&0.5393$\pm$\small{0.021}&0.5014$\pm$\small{0.026}&0.2552$\pm$\small{0.007}&1.2666$\pm$\small{0.014}&\textbf{1.0}$\pm$\small{0.0}&1.3814$\pm$\small{0.001}\\

\cmidrule{2-12}
&$+\mathcal{D}$  &DNN&33.1131$\pm$\small{0.097}&80.1763$\pm$\small{0.238}&&0.5389$\pm$\small{0.116}&0.3929$\pm$\small{0.057}&0.2867$\pm$\small{0.037}&1.2548$\pm$\small{0.06}&0.9831$\pm$\small{0.018}&1.38$\pm$\small{0.003}\\
&$+\mathcal{D}$  &DC&\underline{40.3372}$\pm$\small{0.07}&89.1184$\pm$\small{0.039}&&\underline{0.2649}$\pm$\small{0.127}&\underline{0.2576}$\pm$\small{0.105}&\underline{0.1568}$\pm$\small{0.058}&\underline{1.0539}$\pm$\small{0.111}&\underline{0.9997}$\pm$\small{0.001}&\underline{1.3496}$\pm$\small{0.021}\\
&$+\mathcal{D}$  &MCD&34.1571$\pm$\small{0.029}&\underline{89.1436}$\pm$\small{0.026}&&0.5403$\pm$\small{0.043}&0.5074$\pm$\small{0.015}&0.2663$\pm$\small{0.013}&1.2694$\pm$\small{0.026}&\textbf{1.0}$\pm$\small{0.0}&1.3821$\pm$\small{0.002}\\
&$+\mathcal{D}$  &VI&\textbf{41.8279}$\pm$\small{0.073}&\textbf{91.0999}$\pm$\small{0.015}&&\textbf{0.1634}$\pm$\small{0.051}&\textbf{0.1403}$\pm$\small{0.064}&\textbf{0.0861}$\pm$\small{0.029}&\textbf{0.9464}$\pm$\small{0.066}&0.9958$\pm$\small{0.004}&\textbf{1.3246}$\pm$\small{0.019}\\

\midrule
\parbox[t]{2mm}{\multirow{8}{*}{\rotatebox[origin=c]{90}{MEDMCQA \resizebox{2ex}{!}{\worldflag{FR}}}}}
& $-\mathcal{D}$&DNN&28.5727$\pm$\small{0.03}&\underline{63.88}$\pm$\small{0.055}&&0.6787$\pm$\small{0.1}&0.3256$\pm$\small{0.043}&0.1575$\pm$\small{0.021}&1.5347$\pm$\small{0.062}&0.9625$\pm$\small{0.033}&1.6063$\pm$\small{0.003}\\
& $-\mathcal{D}$&DC&\textbf{32.0291}$\pm$\small{0.003}&63.5584$\pm$\small{0.007}&&\textbf{0.4822}$\pm$\small{0.015}&\textbf{0.165}$\pm$\small{0.01}&\textbf{0.1099}$\pm$\small{0.0}&\textbf{1.3846}$\pm$\small{0.009}&0.9764$\pm$\small{0.007}&\textbf{1.5888}$\pm$\small{0.001}\\
& $-\mathcal{D}$&MCD&28.3648$\pm$\small{0.029}&61.3612$\pm$\small{0.103}&&0.7533$\pm$\small{0.044}&0.3819$\pm$\small{0.084}&0.1518$\pm$\small{0.02}&1.5848$\pm$\small{0.024}&\textbf{1.0}$\pm$\small{0.0}&1.6091$\pm$\small{0.0}\\
& $-\mathcal{D}$&VI&23.1977$\pm$\small{0.042}&48.5531$\pm$\small{0.046}&&0.7499$\pm$\small{0.023}&0.242$\pm$\small{0.033}&0.1329$\pm$\small{0.004}&1.5822$\pm$\small{0.013}&\textbf{1.0}$\pm$\small{0.0}&1.6089$\pm$\small{0.0}\\
\cmidrule{2-12}
&$+\mathcal{D}$  &DNN&28.1549$\pm$\small{0.045}&61.0932$\pm$\small{0.089}&&0.6859$\pm$\small{0.12}&0.3026$\pm$\small{0.009}&0.1582$\pm$\small{0.01}&1.5388$\pm$\small{0.077}&0.9775$\pm$\small{0.02}&1.6064$\pm$\small{0.004}\\
&$+\mathcal{D}$  &DC&29.7558$\pm$\small{0.07}&60.343$\pm$\small{0.103}&&0.6687$\pm$\small{0.17}&0.2973$\pm$\small{0.069}&0.1278$\pm$\small{0.018}&1.5216$\pm$\small{0.122}&0.9893$\pm$\small{0.019}&1.6025$\pm$\small{0.012}\\
&$+\mathcal{D}$  &MCD&\underline{31.0912}$\pm$\small{0.016}&\textbf{68.4352}$\pm$\small{0.033}&&\underline{0.5541}$\pm$\small{0.115}&0.3122$\pm$\small{0.059}&0.1477$\pm$\small{0.031}&\underline{1.4543}$\pm$\small{0.081}&\underline{0.9936}$\pm$\small{0.011}&\underline{1.5999}$\pm$\small{0.007}\\
&$+\mathcal{D}$  &VI&23.1243$\pm$\small{0.035}&49.8553$\pm$\small{0.031}&&0.7415$\pm$\small{0.017}&\underline{0.2336}$\pm$\small{0.026}&\underline{0.1222}$\pm$\small{0.008}&1.5765$\pm$\small{0.01}&\textbf{1.0}$\pm$\small{0.0}&1.6085$\pm$\small{0.0}\\

\midrule
\parbox[t]{2mm}{\multirow{8}{*}{\rotatebox[origin=c]{90}{MORFITT \resizebox{2ex}{!}{\worldflag{FR}}}}}

& $-\mathcal{D}$&DNN&49.7506$\pm$\small{0.009}&59.038$\pm$\small{0.012}&&0.6499$\pm$\small{0.022}&0.2323$\pm$\small{0.021}&0.0398$\pm$\small{0.005}&2.0748$\pm$\small{0.015}&0.796$\pm$\small{0.045}&2.4454$\pm$\small{0.003}\\
& $-\mathcal{D}$&DC&\underline{55.4551}$\pm$\small{0.01}&\underline{62.5306}$\pm$\small{0.008}&&0.6134$\pm$\small{0.003}&0.2243$\pm$\small{0.003}&0.0425$\pm$\small{0.001}&2.0332$\pm$\small{0.006}&0.8775$\pm$\small{0.014}&2.4411$\pm$\small{0.001}\\
& $-\mathcal{D}$&MCD&48.3269$\pm$\small{0.008}&57.3529$\pm$\small{0.008}&&0.6309$\pm$\small{0.021}&0.1519$\pm$\small{0.05}&0.0464$\pm$\small{0.007}&2.2692$\pm$\small{0.03}&\textbf{0.9856}$\pm$\small{0.006}&2.4767$\pm$\small{0.003}\\
& $-\mathcal{D}$&VI&53.0834$\pm$\small{0.014}&61.6728$\pm$\small{0.01}&&0.6408$\pm$\small{0.042}&0.2571$\pm$\small{0.039}&0.0477$\pm$\small{0.006}&\textbf{2.0245}$\pm$\small{0.007}&0.7724$\pm$\small{0.047}&\textbf{2.4369}$\pm$\small{0.004}\\
\cmidrule{2-12}
&$+\mathcal{D}$  &DNN&53.4963$\pm$\small{0.019}&61.8015$\pm$\small{0.014}&&0.6081$\pm$\small{0.017}&0.2098$\pm$\small{0.014}&0.0363$\pm$\small{0.002}&2.0538$\pm$\small{0.015}&0.8334$\pm$\small{0.01}&2.4453$\pm$\small{0.002}\\
&$+\mathcal{D}$  &DC&\textbf{56.4418}$\pm$\small{0.018}&\textbf{62.9596}$\pm$\small{0.02}&&0.6148$\pm$\small{0.027}&0.2325$\pm$\small{0.018}&0.0433$\pm$\small{0.003}&\underline{2.0251}$\pm$\small{0.015}&0.8667$\pm$\small{0.03}&\underline{2.4394}$\pm$\small{0.001}\\
&$+\mathcal{D}$  &MCD&51.8519$\pm$\small{0.015}&60.5392$\pm$\small{0.006}&&\underline{0.5718}$\pm$\small{0.003}&\underline{0.0687}$\pm$\small{0.022}&\underline{0.0298}$\pm$\small{0.0}&2.1426$\pm$\small{0.01}&0.9651$\pm$\small{0.005}&2.4629$\pm$\small{0.002}\\
&$+\mathcal{D}$  &VI&54.2993$\pm$\small{0.011}&62.7145$\pm$\small{0.01}&&\textbf{0.5346}$\pm$\small{0.008}&\textbf{0.0488}$\pm$\small{0.018}&\textbf{0.0279}$\pm$\small{0.002}&2.1064$\pm$\small{0.014}&\underline{0.9752}$\pm$\small{0.007}&2.4602$\pm$\small{0.002}\\
         \bottomrule
    \end{tabular}
    }
    \caption{Comparison for text classification performance and uncertainty-awareness. We report the mean of 10 seed runs for all the metrics. We denote best score with \textbf{bold} and second best with \underline{underline}. We denote both English and French domain-specific PLMs with $+\mathcal{D}$. The models DC, MCD, VI are from the $+\mathcal{U}$ set.} 
    \label{tab:final-all-seeds}
\end{table*}

\section{Full results}
\label{adx:sup results}
We present the detailed Table for all the configurations in Table \ref{tab:final-all-seeds}.
As noted in the main text, the most obvious trend across the board is that scores are tightly coupled with datasets: 
The range of scores achieved by all classifiers we study tends to be fairly limited across a given dataset, whereas we can observe often spectacular differences from one dataset to the next.

Insofar as classification metrics go, we observe that $+\mathcal{D}$ models almost always occupy the top ranks.
This is especially salient in MedABS and MedNLI, where all $+\mathcal{D}$ classifiers outperform all $-\mathcal{D}$ classifiers both in terms of F1 and accuracy.
In PxSLU, the only model that deviates from this trend is the $+\mathcal{D}-\mathcal{U}$ model, which appears to suffer from an especially low accuracy.
In the two other French datasets, along with SMOKING, classification metrics do not exhibit as clear a division between domain-specific and general PLMs.

As for calibration metrics, we find a very similar behavior to what we highlight in the main text: uncertainty-unaware model almost never rank among the top two contenders.
Rankings per metric tend to be fairly stable as long as we control for domain-specificity.

Lastly, having a look at the various Bayesian architecture, we can see that DropConnect is not necessarily the most optimal system across all uncertainty-aware classifiers.
Selecting the best architectures given 3 seeds, and then expanding to 10 seeds most likely led to some degree of sampling bias, explaining this discrepancy.
It does however constitute a strong contender across many situations: it still remains the best ranking Bayesian architecture on average both in terms of F1 across the validation set, as well as in terms of test BS., ECE, SCE, NLL and Entropy.

In fact, differences in terms of ranks across datasets per architecture are not always significant: If we normalize all 80 classifiers per dataset by taking their rank, then Kruskal-Wallis H-test suggest that F1, accuracy and ECE do not lead to significant rank differences across architectures (assuming a threshold of $p < 0.05$).
Likewise, comparing $+\mathcal{D}$ and $-\mathcal{D}$ models with the same procedure does not lead to significant differences in terms of ECE, SCE, and coverage.



\end{document}